\documentclass[journal,twoside,web]{ieeecolor}
\usepackage{generic}
\usepackage{cite}
\usepackage{amsmath,amssymb,amsfonts}
\usepackage{mathrsfs}
\usepackage{algorithm}
\usepackage{algorithmicx}
\usepackage{algpseudocode}
\usepackage{graphicx}
\usepackage{textcomp}
\usepackage{hyperref}
\hypersetup{hidelinks=true}
\usepackage{multirow}
\usepackage{booktabs}
\usepackage{adjustbox}
\usepackage{xcolor}
\usepackage{soul}
\usepackage{listings}
\usepackage[utf8]{inputenc}
\usepackage{manyfoot}

\def\BibTeX{{\rm B\kern-.05em{\sc i\kern-.025em b}\kern-.08em
    T\kern-.1667em\lower.7ex\hbox{E}\kern-.125emX}}

\begin{document}

\title{HMSViT: A Hierarchical Masked Self-Supervised Vision Transformer for Efficient and Robust Corneal Nerve Segmentation and Diabetic Neuropathy Diagnosis}

\author{Xin Zhang, Liangxiu Han*, Yue Shi, Tam Sobeih, Yalin Zheng, Uazman Alam, Maryam Ferdousi, and Rayaz Malik
\thanks{This work was supported by the Engineering and Physical Sciences Research Council (EPSRC) through grants EP/X013707/1 and EP/X01441X/1. (Corresponding author: Liangxiu Han.)}
\thanks{X. Zhang, L. Han*, Y. Shi, and T. Sobeih are with the Department of Computing and Mathematics, Manchester Metropolitan University, Manchester M15GD, UK (e-mail: x.zhang@mmu.ac.uk; l.han@mmu.ac.uk; y.shi@mmu.ac.uk; T.Sobeih@mmu.ac.uk).}
\thanks{Y. Zheng and U. Alam are with the Department of Eye and Vision Sciences, University of Liverpool, Liverpool L78TX, UK (e-mail: yalin.zheng@liverpool.ac.uk; uazman.alam@liverpool.ac.uk).}
\thanks{M. Ferdousi is with the Faculty of Biology, Medicine and Health, University of Manchester, Manchester, UK (e-mail: maryam.ferdousi@manchester.ac.uk).}
\thanks{R. Malik is with the Department of Medicine, Weill Cornell Medicine-Qatar, Doha, Qatar (e-mail: ram2045@qatar-med.cornell.edu).}
}

\maketitle

\begin{abstract}
Diabetic peripheral neuropathy (DPN) is a common and disabling complication of diabetes, necessitating early and accurate detection. Corneal confocal microscopy (CCM) offers a non-invasive diagnostic window, but automated analysis is often limited by inefficient feature extraction and scarce annotations. We propose HMSViT, a hierarchical masked self-supervised vision transformer for corneal nerve segmentation and DPN diagnosis. HMSViT uses a pooling-based hierarchical architecture with a dual-attention design and absolute positional encoding to capture multi-scale features efficiently. The model is pretrained with block-masked self-supervised learning to improve robustness and reduce reliance on labelled data, and a lightweight multi-scale decoder is used for downstream segmentation and classification. Experiments on clinical CCM datasets using patient-level 5-fold cross-validation demonstrate strong performance, achieving 85.6\% patient-level diagnostic accuracy and 61.34\% mIoU for nerve segmentation. Compared with Swin Transformer and HiViT, HMSViT improves segmentation mIoU by 2.45--3.04\% and diagnostic accuracy by 2.3--3.8\% while using up to 41\% fewer parameters. Ablation studies show that combining hierarchical multi-scale design with block-masked pretraining is critical for optimal performance.
\end{abstract}


\section{Introduction}
\label{sec:introduction}
Diabetic peripheral neuropathy (DPN) is a debilitating complication of diabetes, affecting nearly 50\% of individuals with diabetes worldwide and significantly increasing the risk of foot ulcers, infections, and lower-limb amputations \cite{Iqbal2018Diabetic}. This progressive neurodegenerative condition is characterised by nerve damage, primarily in the peripheral nervous system, leading to symptoms such as pain, numbness, and loss of sensation in the extremities. Early diagnosis and continuous monitoring are essential to mitigate its impact and prevent irreversible nerve damage. 

However, traditional diagnostic methods, including nerve conduction studies and skin biopsies, are invasive, costly, and often unsuitable for routine screening \cite{Burgess2021Early}. In recent years, Corneal Confocal Microscopy (CCM) has emerged as a non-invasive and highly sensitive imaging modality for early DPN detection \cite{Chen2016automatic}. CCM enables high-resolution imaging of corneal nerve fibres, which serve as quantifiable biomarkers for neuropathy progression. Despite its potential, the manual analysis of CCM images is time-consuming, prone to inter-observer variability, and requires expert interpretation, making it impractical for large-scale screening \cite{Gad2022Corneal,Petropoulos2021Corneal}. These challenges underscore the need for automated, AI-driven solutions that can efficiently analyse CCM images for both corneal nerve segmentation and DPN severity classification.

The advent of deep learning has significantly advanced medical image analysis, with Convolutional Neural Networks (CNNs) being widely adopted for automated segmentation and classification \cite{Sarvamangala2022Convolutional,Williams2020artificial}. While CNNs are exceptional at extracting hierarchical feature representations, their fixed local receptive fields can hinder the capture of long-range dependencies and global structural variations—an essential factor in accurately interpreting complex nerve fibre morphology \cite{Williams2020artificial}. To overcome these constraints, Vision Transformers (ViTs) \cite{Dosovitskiy2020Image}  have emerged as a potent alternative, employing self-attention mechanisms to harness global contextual information, which is particularly advantageous for medical imaging tasks requiring fine-grained structural recognition \cite{Azad2024Advances}.

Nonetheless, standard ViTs face notable challenges in medical imaging, especially for tasks demanding high-resolution representations, such as organ segmentation or nerve fibre analysis \cite{Azad2024Advances,Zhou2021review}. Firstly, ViTs require extensive annotated datasets, which are often unavailable in medical contexts, and their patch-based processing reduces spatial resolution and lacks inductive bias for spatial reasoning - the ability to understand structure and relationships within an image. These limitations call for innovative solutions to harness ViTs’ strengths while addressing their shortcomings.

Prior studies have explored hierarchical Vision Transformer architectures, such as the Swin Transformer\cite{Liu2021Swin} and HiViT\cite{Zhang2022HiViT} that progressively merge patches to extract multi-scale features, thereby capturing both local detail and global context. However, these methods often depend on handcrafted modules (e.g., shifted windows or convolutional enhancements) that increase computational complexity. In parallel, self-supervised learning (SSL) techniques, notably masked autoencoders, have been applied to alleviate the dependency on large labelled datasets, although integrating SSL with hierarchical models remains challenging.

To address these issues, we propose HMSViT—a Hierarchical Masked Self-Supervised Vision Transformer. Unlike existing hierarchical ViTs that rely on shifted windows or convolutional priors, HMSViT introduces a pooling-based hierarchical design with dual attention mechanisms, where early layers capture fine-grained local details and deeper layers integrate global contextual information using absolute positional encodings. A novel block-masked SSL approach is introduced to learn robust spatial relationships from unlabelled CCM images, reducing reliance on extensive annotations. A multi-scale decoder further fuses these features for both accurate corneal nerve segmentation and DPN diagnosis. In summary, the key contributions are:

\begin{enumerate}
\item We propose HMSViT, a novel hierarchical Vision Transformer architecture designed specifically for medical image analysis, which integrates dual attention mechanisms and pooling-based token aggregation to efficiently capture both fine-grained local details and global contextual information from high-resolution CCM images.
\item We introduce a novel block-masked self-supervised learning strategy tailored for hierarchical vision transformers, enabling robust feature representation from unlabelled data and significantly reducing reliance on extensive expert annotations—a key barrier in clinical AI deployment.
\item We demonstrate that HMSViT achieves state-of-the-art performance on real-world clinical datasets, outperforming leading models such as Swin Transformer and HiViT in both segmentation and diagnosis, while using fewer parameters and offering strong potential for scalable clinical deployment.
\end{enumerate}

\section{Related work}

\subsection{Deep Learning Methods for Corneal Nerve Segmentation and DPN Diagnosis}

Diabetes affects 463 million people worldwide, projected to reach 700 million by 2045 \cite{Saeedi2019Global}. DPN, impacting up to 50\% of patients, causes 50--75\% of non-traumatic lower-limb amputations \cite{Bodman2025Diabetic}. Defined as peripheral nerve dysfunction in diabetes after excluding other causes \cite{Boulton2005Diabetic}, DPN highlights the urgent need for early detection to prevent irreversible damage.

Corneal Confocal Microscopy (CCM) is a non-invasive tool for early DPN detection, allowing quantitative analysis of the corneal sub-basal nerve plexus, which is highly sensitive to metabolic damage \cite{ball2015improving}. Manual annotation of CCM images is time-consuming, expertise-dependent, and subject to variability due to noise and interpretation bias \cite{Petropoulos2021Corneal}. These limitations have led to the development of automated methods to extract key diagnostic metrics, including corneal nerve fibre length (CNFL), density (CNFD), and branch density (CNBD). Recent advancements in deep learning have significantly improved automated CCM image analysis, particularly in the areas of corneal nerve segmentation and DPN classification. The most prominent approaches include Convolutional Neural Networks (CNNs) and ViTs, both of which have demonstrated success in various medical imaging applications.

\subsubsection{CNN-Based Approaches}

CNNs have been extensively used for DPN classification and corneal nerve segmentation, leveraging their ability to capture hierarchical spatial features. Several CNN-based approaches have been explored in recent years. \cite{Preston2022Artificial} developed a CNN-based classifier for DPN detection using CCM images, incorporating data augmentation techniques to improve model generalisation. Similarly, \cite{Meng2023Artificial} applied ResNet architectures, achieving high classification accuracy in identifying neuropathic changes. For segmentation, \cite{Qiao2024Deep} employed U2Net to precisely quantify corneal nerve fibre parameters, significantly enhancing the sensitivity of DPN detection. \cite{Williams2020artificial} further demonstrated that CNN-based segmentation approaches could outperform traditional corneal nerve analysis software (e.g., ACC Metrics), providing more accurate and reliable quantification of nerve fibre morphology. In addition, \cite{TEH2022455} implemented a 3D CNN model to classify painful DPN subtypes, highlighting the potential of CNNs in personalised medicine and treatment response analysis.

Despite their effectiveness, CNN-based approaches are fundamentally constrained by fixed local receptive fields, limiting their ability to model long-range dependencies and complex nerve fibre branching patterns. Since CNNs rely on localized feature extraction, they struggle to capture the global structure of corneal nerve morphology, which is critical for accurate segmentation. Additionally, CNN-based models are highly sensitive to image contrast variations and noise, making their performance inconsistent across different datasets. These limitations have led researchers to explore ViTs, which leverage self-attention mechanisms to capture both local and global spatial dependencies.

\subsubsection{Vision Transformers (ViTs)} 

ViTs have gained increasing attention in medical imaging due to their ability to model global spatial relationships. Unlike CNNs, which process images through localized convolutional operations, ViTs divide images into patches and apply self-attention mechanisms across the entire image, enabling them to capture long-range dependencies \cite{Azad2024Advances}.This global view is critical for accurately identifying and segmenting nerve fibres that often extend over large parts of CCM images \cite{Dosovitskiy2020Image}.

Several studies have explored ViT-based models for corneal nerve segmentation and DPN classification.  \cite{li2022context} introduced a ViT-based context encoder that integrates channel-wise attention mechanisms, improving segmentation accuracy by preserving subtle nerve fibre structures. Similarly, \cite{Chen2024Development} applied a transformer-based classifier for DPN detection, demonstrating higher diagnostic accuracy compared to CNN-based approaches. While ViTs offer advantages in modeling global context, they suffer from limitations. The patch-based tokenisation in ViTs results in spatial resolution loss, making them less effective for dense segmentation tasks such as nerve fibre analysis.

\begin{table*}[!htbp]
    \centering
    \begin{adjustbox}{max width=\textwidth}
        \begin{tabular}{p{3cm}p{6cm}p{7cm}}
            \hline
            & \textbf{Standard Vision Transformer (ViT)} & \textbf{Hierarchical Vision Transformer (HVT)} \\ \hline
            Patch Handling & Fixed-size patches across all layers & Progressive patch merging \& downsampling \\
            Computational Complexity & Quadratic in image resolution & Reduced complexity with hierarchical tokenisation  \\
            Feature Extraction & Uniform token processing & Multi-scale feature learning (local to global) \\
            Examples & Original ViT\cite{Dosovitskiy2020Image}, DeiT \cite{Touvron2021Training}, DPT \cite{Ranftl2021Vision} & Swin Transformer\cite{Liu2021Swin}, PVT \cite{Wang2021Pyramid}, HiViT\cite{Zhang2022HiViT} \\
            Performance & Performs well on image classification, especially with large datasets and pretraining. & Often achieves better performance on dense prediction tasks like object detection and semantic segmentation due to multi-scale feature representation. \\ \hline
        \end{tabular}
    \end{adjustbox}
    \caption{Comparison between Standard Vision Transformer (ViT) and Hierarchical Vision Transformer (HVT) architectures, highlighting differences in patch handling, computational complexity, feature extraction, representative models, and typical performance characteristics.}
    \label{tab:vit}
\end{table*}

To address the limitations of standard ViTs, Hierarchical Vision Transformers (HVTs) introduce multi-scale feature extraction, progressively refining spatial resolution through patch merging and downsampling. Table~\ref{tab:vit} provides a concise comparison between standard ViTs and Hierarchical ViTs. Standard ViTs process images with fixed-size patches across all layers, resulting in uniform tokenisation. While effective, this approach leads to quadratic computational complexity and requires large datasets for training. In contrast, Hierarchical ViTs use progressive patch merging and downsampling, reducing computational cost and enabling multi-scale feature learning. As a result, HVTs efficiently capture both local and global structures and often generalise better on smaller datasets by leveraging spatial hierarchies (see Table~\ref{tab:vit}).

Several hierarchical transformer architectures have been developed for medical image segmentation. The Swin Transformer \cite{Liu2021Swin} introduced shifted window attention, balancing local and global feature representation while reducing computational complexity. The HiViT model \cite{Zhang2022HiViT} refined hierarchical tokenisation , demonstrating that simpler architectures could achieve competitive segmentation performance. The MViTv2 \cite{Li2022MViTv2} framework improved multi-scale learning by integrating decomposed positional embeddings and residual pooling, allowing for more efficient spatial relationship modeling.

Despite these advances, the application of hierarchical HVTs in medical imaging remains challenging. Although existing HVTs offer multi-scale representations, they often rely on explicit spatial priors—such as convolutions \cite{Zhang2022HiViT}, shifted windows \cite{Liu2021Swin}, and complex relative positional encodings to encode locality and spatial bias. While these mechanisms improve supervised accuracy, they also increase computational complexity, slow inference, and reduce architectural flexibility. These drawbacks are particularly problematic in medical contexts, where efficiency and adaptability are critical. Moreover, no prior study has investigated the application of HVTs for CCM image segmentation or DPN diagnosis, leaving a significant gap in clinical applications.

At the same time, the scarcity of annotated datasets limits the effectiveness of purely supervised approaches in medical imaging. This has motivated the adoption of self-supervised learning (SSL), which allows models to exploit abundant unlabelled data to learn transferable representations. Yet, existing hierarchical designs are not well suited to SSL, as their reliance on heavy spatial priors complicates pretraining and reduces adaptability. These challenges highlight the need for new hierarchical methods that are both computationally efficient and naturally compatible with self-supervised learning.

\subsection{Self-Supervised Learning (SSL)}

Self-supervised learning (SSL) has become essential in computer vision, enabling models to learn robust representations from large unlabelled datasets. This approach is particularly valuable for ViTs, which require extensive data to achieve competitive performance compared to convolutional neural networks \cite{Khan2024Survey}. SSL methods rely on pretext tasks such as predicting missing image parts or distinguishing between different augmented views of an image \cite{Albelwi2022Survey}, helping models capture spatial and semantic information without labelled data.

Early SSL methods introduced tasks including context prediction \cite{Doersch2016Unsupervised}, inpainting \cite{Pathak2016Context}, solving jigsaw puzzles \cite{Noroozi2017Unsupervised}, and rotation prediction \cite{Gidaris2018Unsupervised}, laying a foundation for more advanced techniques. Later, contrastive approaches like SimCLR \cite{Chen2020Simple} and MoCo \cite{He2020Momentum} further improved SSL by learning representations that align similar views and separate unrelated images.

Recently, masked autoencoders (MAE) \cite{He2021Masked} advanced SSL for ViTs by masking image patches and training models to reconstruct the missing pixels, efficiently capturing both global and fine-grained image details. Empirical results demonstrate that MAE accelerates training and achieves superior performance in benchmark tasks, including ImageNet classification and dense prediction, surpassing traditional SSL approaches \cite{Ryali2023Hiera}.

Several attempts have been made to integrate masked SSL into hierarchical ViTs. Approaches such as MaskFeat \cite{Wei2023Masked}, MixMAE \cite{Liu2023MixMAE}, and Fast-iTPN \cite{Tian2024Fast} tackle this challenge by processing masked patches using specialised masked tokens. However, these methods often incur substantial computational overhead because a large portion of computation is devoted to masked tokens, which slows down training.

In contrast, we present block masked self-supervised learning for hierarchical ViT pretraining, resulting in a model that is both powerful and efficient. To the best of our knowledge, no existing work jointly integrates hierarchical ViTs with block-level masked self-supervised learning for corneal confocal microscopy–based neuropathy diagnosis.

\section{The Proposed Method}

\begin{figure*}[!htbp]
    \centering
    \includegraphics[width=0.9\textwidth]{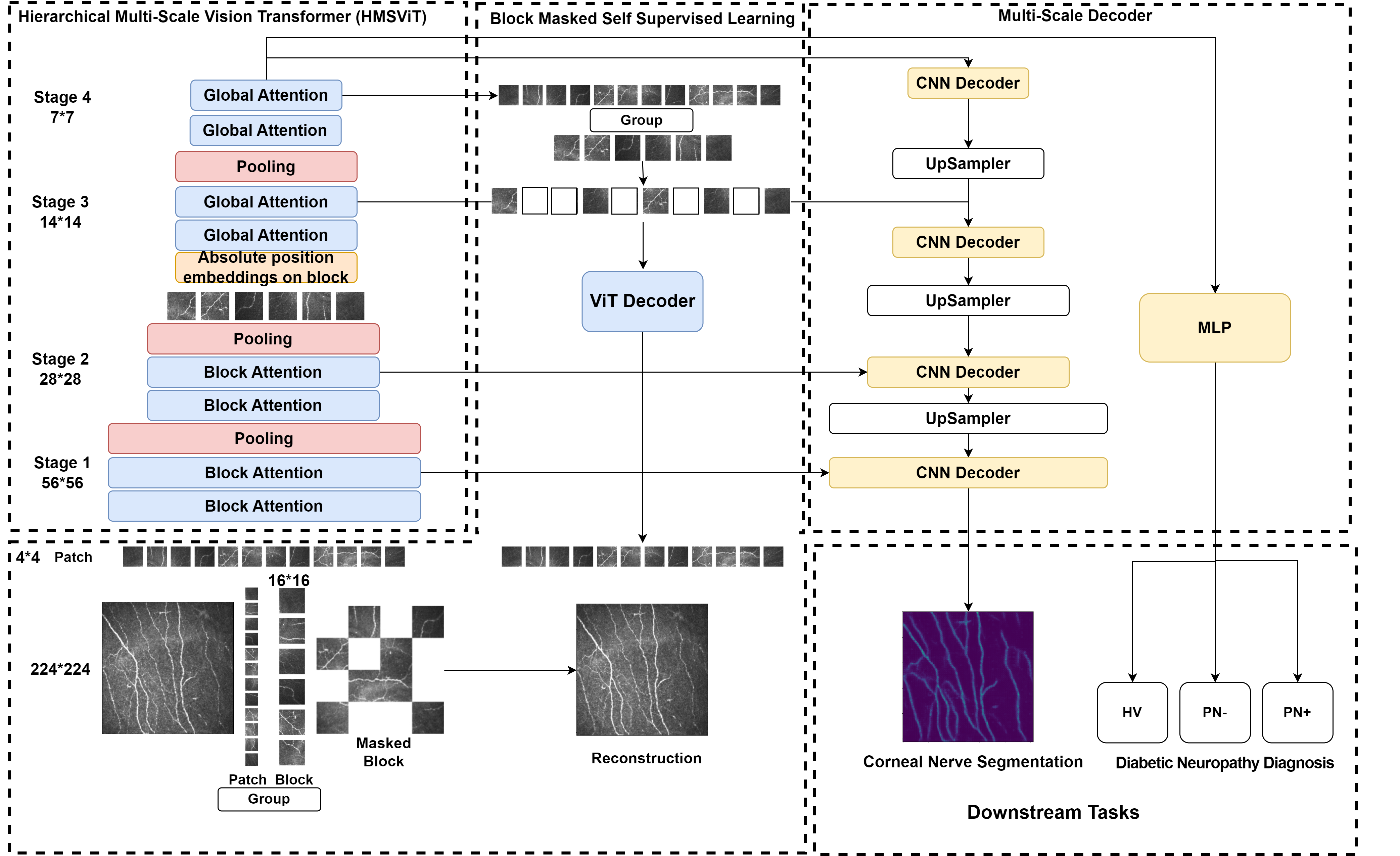}
    \caption{Overview of the proposed method. 1) HMSViT is designed to learn multi-scale representations across four stages through pooling based dual attention mechanisms; 2) A block masked self-supervised learning is used to pretrain the HMSViT to enhance the model's spatial reasoning by reconstructing masked input blocks; 3) A multi-scale decoder is designed for downstream tasks including corneal nerve segmentation and Diabetic neuropathy diagnosis.}
    \label{fig:1}
\end{figure*}

\subsection{Overview of the Proposed Framework}

In this work, we introduce HMSViT—a hierarchical vision transformer for automated corneal nerve segmentation and DPN classification (Figure~\ref{fig:1}). Its design centers on three components:

\begin{enumerate}

\item \textbf{Hierarchical Multi-Scale Feature Extraction:} HMSViT processes images through four hierarchical stages, starting with high-resolution inputs to capture fine details and gradually reducing spatial resolution to extract higher-level features. Instead of relying on complex modules (e.g., shifted windows or convolutions), our design uses a non-parametric pooling operator, with block-based local attention in early stages and global attention in deeper stages complemented by block-level absolute positional encodings.

\item \textbf{Block Masked Self-Supervised Learning:}To learn robust spatial relationships without heavy labelled data, we adapt a Block masked SSL approach. Rather than masking individual patches, we group 4×4 patches into larger 16×16 blocks, aligning with the hierarchical structure and reducing computational overhead.

\item \textbf{Multi-Scale Decoder:} A multi-scale decoder fuses hierarchical features for downstream tasks, enabling effective corneal nerve segmentation and diabetic neuropathy diagnosis.

\end{enumerate}

\subsection{Hierarchical Multi-Scale Vision Transformer (HMSViT)} 

\begin{figure*}[!htbp]
    \centering
    \includegraphics[width=\textwidth]{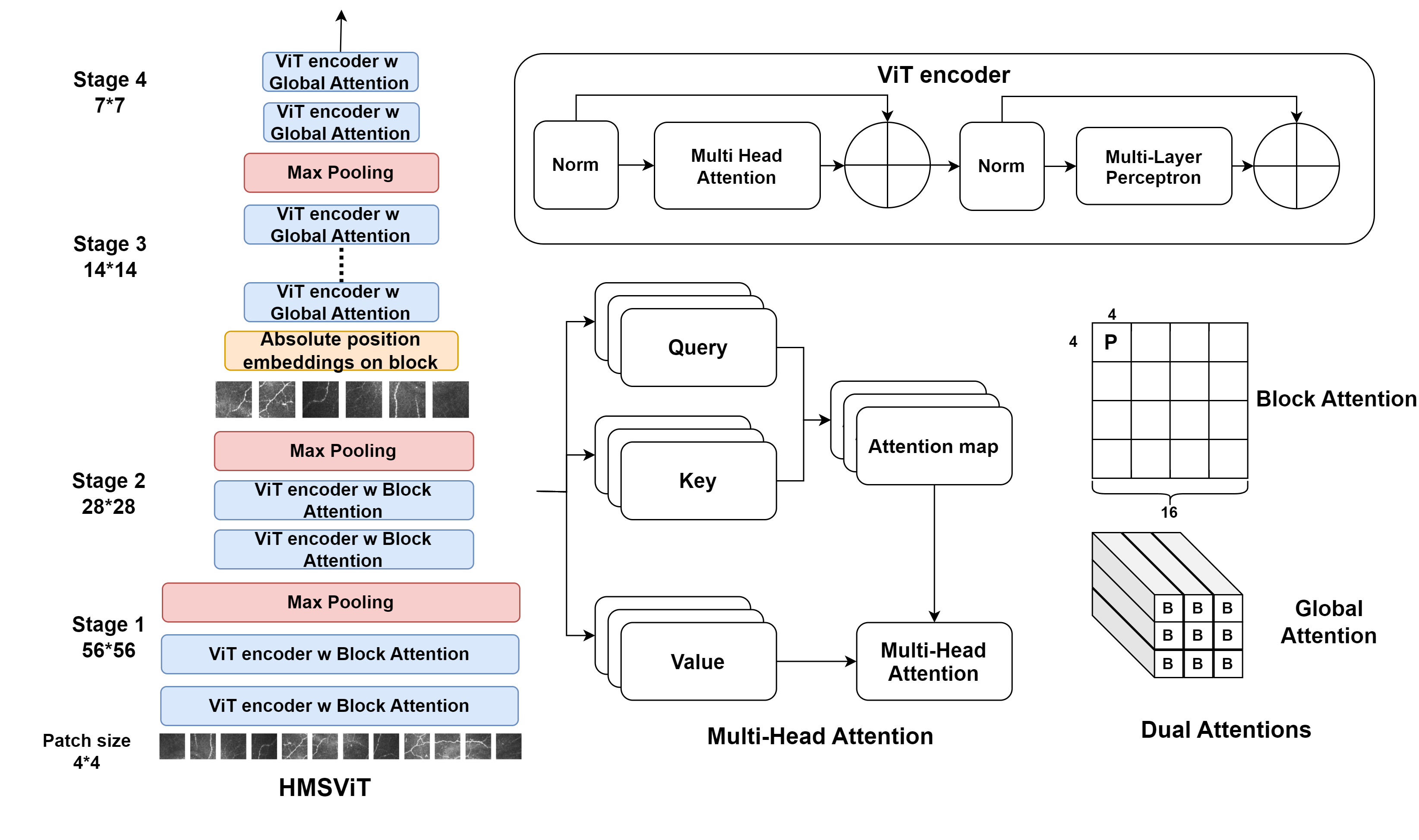}
    \caption{The architecture of the Hierarchical Multi-Scale Vision Transformer (HMSViT). An input image is processed through four hierarchical stages. Early stages use block-based local attention on high-resolution feature maps, while later stages switch to global attention after pooling reduces the spatial dimensions. This dual-attention approach balances computational efficiency with the ability to capture both local details and global context.}
    \label{fig:2}
\end{figure*}

The proposed method leverages the strengths of ViTs while addressing their limitations in dense medical imaging tasks specifically, for corneal nerve fibre segmentation and DPN diagnosis. By progressively extracting features at multiple scales, the network captures both fine-grained local details and broader global context, leading to enhanced segmentation performance and improved diagnostic accuracy (see Figure~\ref{fig:2}).

\subsubsection{High-Resolution Hierarchical Representations}

First, for an input image $X \in \mathbb{R}^{H \times W \times C}$, where $H$ and $W$ are the height and width of the input image and $C$ is the number of channels. HMSViT employs a small patch size of \(4 \times 4\) to obtain high-resolution feature maps. This results in a feature map of size \(56 \times 56\) when using a \(224 \times 224\) input image. Then each patch is flattened and linearly embedded to generate a token representation:

\begin{equation}
    X_p = \text{PatchEmbed}(X) \in \mathbb{R}^{L \times d},
\end{equation}

where $L = \frac{H}{p} \times \frac{W}{p}$ is the number of patches, $p$ is the patch size, and $d$ is the feature dimension. $\frac{H}{p} \times \frac{W}{p}$ is the output feature size.

Since the computational complexity of a ViT scales quadratically with the number of tokens (i.e., \(O(L^2)\)), it becomes essential to reduce the number of tokens to decrease computation and enable multi-scale representations. The patch tokens are first rearranged into a new unit called a Block (B). At a given stage \(l\), the tokens are grouped into blocks:

\begin{equation}
    B^{l}= \text{Concat}\{X_1,X_2,X_3....X_{16} \} \in \mathbb{R}^{4 \times 4 \times d} 
\end{equation}

A non-parametric max pooling operator is then applied to each block to downsample the representation. Max pooling is selected to preserve the fine-grained structural details typical of CCM, where nerve fibers appear as thin, high-intensity structures against a noisy background. Max pooling is theoretically ideal as it retains the strongest activation—representing the nerve signal—within a local block, whereas average pooling tends to smooth out these details, potentially obscuring thin nerve signals. Furthermore, compared to strided convolutions, max pooling is parameter-free, reducing model complexity. The pooling operation is defined as:

\begin{equation}
    B^{(l+1)} = \text{MaxPool}\Bigl( B^{l} \Bigr)
\end{equation}

This operation reduces the spatial resolution and aggregates the most dominant features, facilitating efficient hierarchical feature extraction.

\subsubsection{Dual Attention Mechanisms}

Given that high-resolution features in the early stages result in a large number of tokens ($L$), applying global self-attention becomes computationally intensive due to its $O(L^2)$ complexity. To mitigate this, HMSViT employs a dual attention strategy that adapts to the network's changing hierarchical stages, balancing computational efficiency with feature extraction quality (see Figure~\ref{fig:2}). The standard scaled dot-product attention is defined as:

\begin{equation}
    \text{Attention}(Q, K, V) = \text{softmax} \left( \frac{QK^T}{\sqrt{d_k}} \right) V
\end{equation}

where $Q$, $K$, and $V$ are the query, key, and value matrices, and $d_k$ is the dimension of the keys. Our dual mechanism applies this operation at two different scopes:

\begin{itemize}
    \item \textbf{Block-based Local Attention:} In the early stages of the network (e.g., Stages 1 and 2), where feature maps are large, we apply attention locally within each block of tokens. For a given block $B^l$, the query, key, and value matrices ($Q, K, V$) are derived exclusively from the tokens within that block. This constrains the attention mechanism to focus on local, fine-grained features and significantly reduces computational cost, as the sequence length for each attention calculation is small (e.g., 16 tokens).

    \item \textbf{Global Attention:} In the deeper stages (e.g., Stages 3 and 4), the feature maps have been downsampled, resulting in fewer tokens. At this point, the model switches to global attention. Here, the $Q, K, V$ matrices are derived from the entire sequence of tokens in the feature map. This allows the model to capture long-range dependencies and integrate high-level semantic information across the entire image, which is computationally feasible due to the reduced number of tokens.
\end{itemize}

This hybrid strategy allows HMSViT to efficiently process high-resolution inputs by first capturing local patterns with block-based attention and then modeling global context with global attention in later stages. This ensures a comprehensive feature representation without the prohibitive cost of applying global attention at all layers.

\subsubsection{Absolute positional encodings}

To preserve spatial relationships—crucial in medical imaging—we add positional encodings to the input data. While standard ViTs use absolute or relative embeddings (with hierarchical models often favoring relative ones for consistency across varying resolutions), relative embeddings require modifying the attention mechanism and add computational overhead. In our method, we add learnable absolute positional encodings at the block level:

\begin{equation}
    \tilde{B}^{l} = B^{l} + p^{l}
\end{equation}

where \(p^{l}\) is the learnable absolute positional embedding for token \(l\).

\subsection{Block-masked self-supervised learning}

\begin{figure}[!htbp]
    \centering
    \includegraphics[width=0.5\textwidth]{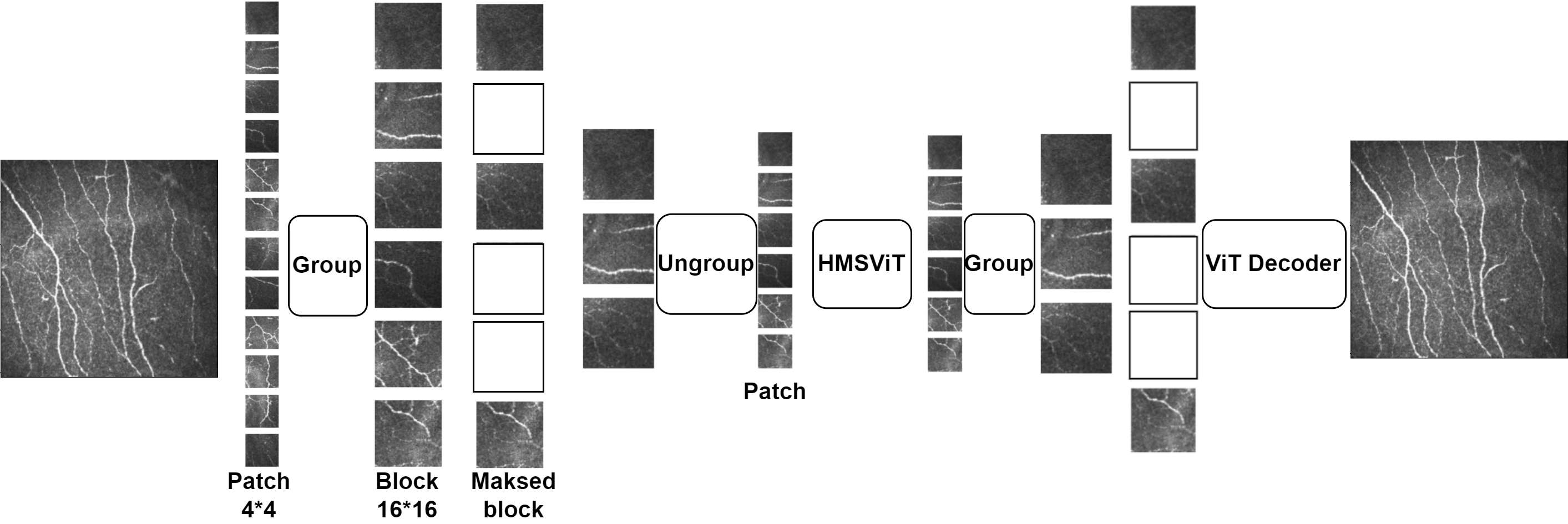}
    \caption{An overview of the block-masked self-supervised learning framework. Patches are grouped into blocks, a portion of which are randomly masked. The HMSViT encoder processes the visible blocks and is trained to reconstruct the original input, forcing it to learn robust spatial representations.}
    \label{fig:3}
\end{figure}

The SSL approach builds on the recent advances in MAE \cite{He2021Masked}, where a large portion of the input (typically 75–90\%) is masked out, and the model is trained to reconstruct the missing parts. This strategy forces the network to learn rich spatial relationships, as it must infer the structure of the masked regions from the visible context. However, in a hierarchical Vision Transformer (ViT) with progressive downsampling, directly applying standard MAE can disrupt spatial coherence across scales. To address this, we propose a block-masking strategy that groups small patches into larger blocks of $16 \times 16$ pixels (comprising $4 \times 4$ patches) before masking. This is motivated by two key factors. First, masking tiny tokens tends to force the model to reconstruct trivial low-level textures, such as edges and noise. By contrast, masking larger blocks helps shift the learning signal from simply "filling in pixel noise" to "understanding the scene", requiring the model to infer higher-level structure. Second, masking larger blocks improves computational efficiency, as it results in fewer mask decisions and fewer visible tokens for the decoder to process. As shown in {Figure~\ref{fig:3}}, we (1) group the patches into block units, (2) randomly mask blocks at the block level, and (3) ungroup them back into smaller patches for our HMSViT.

Given an input feature map $X$, a binary mask $M \in \{0,1\}^{N}$ is generated, where: $X_{\text{unmasked}} = (1-M) \odot X$, $\odot$ denotes element-wise multiplication. Then the visible blocks are fed into HMSViT:

\begin{equation}
    \hat{F} = HMSViT(X_{\text{unmasked}})
\end{equation}
The decoder takes the latent representation (F) and, typically, a set of learnable mask tokens to reconstruct the full input:
\begin{equation}
\hat{X} = Decoder(\hat{F}, \text{masked block tokens})
\end{equation}
The training objective is to minimize the reconstruction error, but only over the masked blocks. The masked autoencoder loss is:
\begin{equation}
    \mathcal{L} = \mathbb{E}_{x \sim p(x)} \left[ \| (1-M) \odot (x - \hat{X}) \|^2 \right]
\end{equation}

where the loss is computed only over the positions where \( M \) indicates a mask (i.e., \(1-M\) selects the masked regions).

\subsection{Multi-scale decoder}

We introduce a multi-scale decoder that fuses HMSViT's hierarchical features for both corneal nerve segmentation and DPN diagnosis.

\begin{figure}[!htbp]
    \centering
    \includegraphics[width=0.5\textwidth]{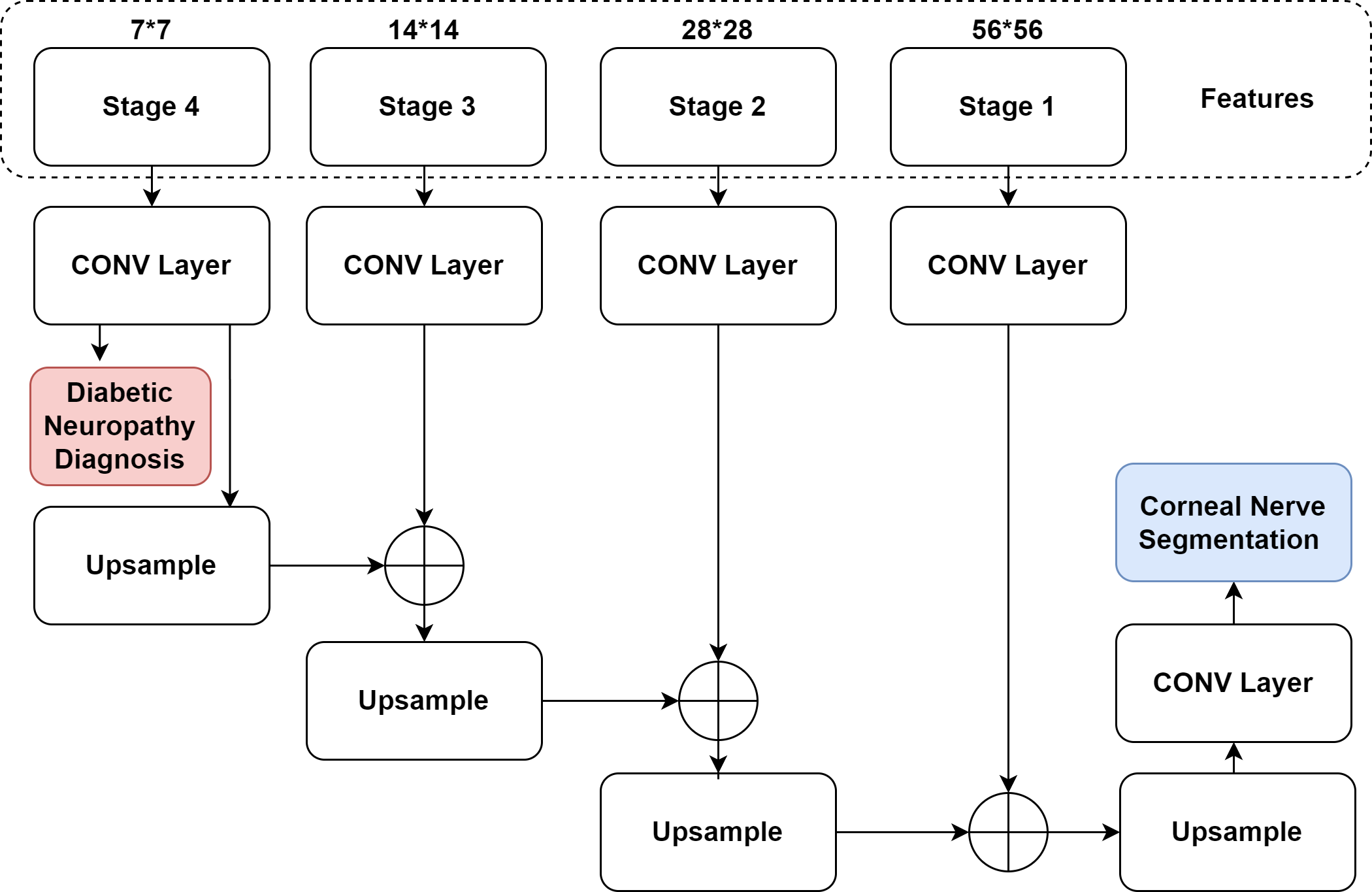}
    \caption{Architecture of the multi-scale decoder. For the segmentation task, hierarchical features from all stages are upsampled and fused to generate a high-resolution segmentation map. For the classification task, the global feature from the final stage is processed by a Multi-Layer Perceptron (MLP) to produce the diagnostic prediction.}
    \label{fig:4}
\end{figure}

For segmentation, as shown in {Figure~\ref{fig:4}}, the hierarchical features \( F_i \) from different stages are fused. Let \( U(\cdot) \) be an upsampling function that brings features to a common resolution. The fused feature map is computed as:

\begin{equation}
    y_{seg}  = \sigma\!\Biggl( \sum_{i} U\bigl(F_i\bigr) \Biggr)
\end{equation}

where \( \sigma(\cdot) \) denotes a convolution with activation functions to refine the aggregated representation. The model is fine-tuned for this task using a combination of Dice loss and Binary Cross-Entropy (BCE) loss to handle class imbalance and produce accurate segmentation maps.

For classification (e.g., DPN diagnosis), the lowest-resolution feature \( F_{\text{global}} \) is fed to an MLP:

\begin{equation}
    y_{classification} = \text{MLP}\Bigl( F_{\text{global}} \Bigr)
\end{equation}

This task is trained using a standard cross-entropy loss function to optimise classification accuracy. These task-specific loss functions are applied only during the fine-tuning stage, separately from the reconstruction loss used during self-supervised pretraining.

\section{Experimental Evaluation}

\subsection{Experimental Design}

The experiments in this study are designed to comprehensively evaluate the effectiveness and efficiency of our proposed self-supervised HMSViT. Three main experiments are conducted: 1) Model Performance Evaluation, where we assess the model on DPN diagnosis and corneal nerve fibre segmentation; 2) Ablation Study, examining the impact of the hierarchical design and self-supervised learning; and 3) Comparison with State-of-the-Art Methods, benchmarking our segmentation results against existing techniques. Each of these tasks is discussed in detail below.

\subsubsection{Task 1: Model Performance Evaluation}

We evaluate HMSViT on two tasks—DPN diagnosis and corneal nerve fibre segmentation—by comparing three configurations (Tiny, Small, Base) with state-of-the-art hierarchical ViT based models (Swin Transformer \cite{Liu2021Swin} and HiViT \cite{Zhang2022HiViT}).

\begin{enumerate}

\item \textbf{DPN Diagnosis Task:} This task assesses the models' ability to classify participants into three categories: Healthy Volunteers (HV), Diabetic Patients without Peripheral Neuropathy ($\text{PN}-$), and Diabetic Patients with Peripheral Neuropathy ($\text{PN}+$). To ensure a comprehensive evaluation, performance is measured at two levels:
\begin{itemize}[leftmargin=*, topsep=0pt, partopsep=0pt, itemsep=0pt]
    \item \textbf{Patient-Level Diagnosis:} To mitigate intra-subject variability and improve diagnostic robustness, predictions from the 4–8 images per participant are aggregated via majority voting \cite{kalteniece2017corneal}. Majority voting reflects clinical practice, where diagnosis is made per patient rather than per image. The resulting patient-level accuracy serves as the primary metric for comparing HMSViT against other state-of-the-art hierarchical models.
    \item \textbf{Image-Level Diagnosis:} To evaluate the intrinsic classification capability of our model, we also report image-level metrics—including precision, recall, F1-score, and accuracy—for each of the three HMSViT configurations (Tiny, Small, and Base).
\end{itemize}

\item \textbf{Corneal Nerve fibre Segmentation Task:} This task evaluates the models' ability to accurately identify and delineate corneal nerve fibres. The primary metric is the mean Intersection over Union (mIoU), which measures the overlap between predicted and actual segmentation regions. In addition, clinically relevant metrics (Table~\ref{tab:task})—Corneal Nerve fibre Length (CNFL) and Corneal Nerve Branch Density (CNBD)—are used to quantify nerve morphology, offering further insights into early DPN detection. The results underscore HMSViT’s superior segmentation accuracy compared to existing hierarchical vision transformers.

\end{enumerate}

\begin{table}[!htbp]
    \centering
    \begin{adjustbox}{max width=\columnwidth}
    \begin{tabular}{lll}
    \hline
    Parameter                             & Description                                    & Unit of Measurement \\ \hline
    Corneal nerve fibre length (CNFL)   & Length of all main nerve fibres and branches & mm/mm²              \\
    Corneal nerve branch density (CNBD) & Number of main nerve fibre branches          & no/mm²              \\ \hline
    \end{tabular}
    \end{adjustbox}
    \caption{Parameters for corneal nerve fibre analysis.}
    \label{tab:task}
\end{table}

\subsubsection{Task 2: Ablation Study to Evaluate the Impact of Hierarchical Design and Self-Supervised Learning}

An ablation study is conducted to assess the impact of key components of the proposed model: 1) Hierarchical Design and 2) Self-Supervised Learning. The experiment is divided into four settings: (1) supervised training without the hierarchical structure, (2) supervised training with the hierarchical structure, (3) SSL without the hierarchical structure, and (4) SSL with the hierarchical structure.

\subsubsection{Task 3: Comparison with State-of-the-Art Methods for Segmentation}

To comprehensively evaluate HMSViT’s segmentation performance, we benchmark it against several state-of-the-art Vision Transformer-based models, including baseline ViTs, hierarchical ViTs, and self-supervised learning (SSL) approaches. These existing methods are summarised in Table~\ref{tab:models}. Our evaluation spans
diverse architectural paradigms, ranging from vanilla transformers to hierarchical
and convolutional variants, ensuring a thorough comparison of model
effectiveness.

\begin{table*}[ht]
    \centering
    \begin{adjustbox}{max width=\textwidth}
    \begin{tabular}{llp{11cm}}
    \hline
    \textbf{Category} & \textbf{Method} & \textbf{Description} \\
    \hline
    Baseline Vision Transformer & DPT \cite{Ranftl2021Vision} & A ViT-based model designed specifically for dense prediction tasks like segmentation. \\
    \hline
    \multirow{4}{*}{\parbox{2.5cm}{Hierarchical Vision Transformers}} 
      & HiViT \cite{Zhang2022HiViT} & Extends ViT by incorporating multi-scale feature representations. \\ \cline{2-3}
      & PoolFormer \cite{Yu2022MetaFormer} & Replaces traditional self-attention with pooling operations to efficiently capture contextual information. \\ \cline{2-3}
      & SegFormer \cite{Xie2021SegFormer} & A transformer optimised for semantic segmentation, with a focus on efficient multi-scale feature aggregation. \\ \cline{2-3}
      & Fast-iTPN \cite{Tian2024Fast} & A pyramid-based architecture that enhances efficiency and downstream performance by integrating token migration and gathering while jointly pretraining both the backbone and the neck. \\
    \hline
    \multirow{3}{*}{\parbox{2.5cm}{Self-Supervised Learning Approaches}} 
      & MoCo v3 \cite{Chen2021Empirical} & Utilises contrastive learning to learn robust visual representations. \\ \cline{2-3}
      & MAE \cite{He2021Masked} & Implements a masked autoencoder strategy to reconstruct missing regions, improving feature learning. \\ \cline{2-3}
      & MixMIM \cite{Liu2023MixMAE} & Enhances MAE by introducing a mixing strategy, replacing masked tokens with tokens from another image to encourage richer feature disentanglement. \\
    \hline
    \end{tabular}
    \end{adjustbox}
    \caption{Summary of representative Vision Transformer-based models and self-supervised learning approaches used for medical image segmentation. The models are classified into baseline ViTs, hierarchical architectures, and self-supervised frameworks, highlighting their core design strategies and contributions to efficient feature learning.}
    \label{tab:models}
\end{table*}

For a comprehensive performance assessment, we use evaluation metrics including mean Intersection over Union (mIoU), a widely adopted metric in segmentation tasks. In this experiment, we focus exclusively on corneal nerve segmentation results because precise nerve delineation is fundamental to both clinical assessment and the overall performance of our system. Accurate segmentation directly impacts the quantification of clinically relevant parameters—such as nerve fibre length and branch density—which are critical for diagnosing DPN. Moreover, most of the hierarchical models we compare are specifically designed for segmentation tasks, ensuring that our evaluation is both fair and directly aligned with the core functionality of these architectures.

\subsection{Datasets}

Our study leverages two distinct data sources to train and evaluate HMSViT: a public dataset and a private clinical cohort. The first is the public CORN CCM dataset \cite{imed_ningbo}, an aggregation of four subsets designed for various tasks including nerve segmentation, image enhancement, and tortuosity grading. This dataset provides a diverse collection of CCM images that enriches our self-supervised pretraining phase. The second is a private clinical UoM CCM dataset from the Early Neuropathy Assessment (ENA) group at the University of Manchester, UK, which includes CCM images from 318 participants. 

These two sources serve distinct but complementary roles: self-supervised pretraining and two downstream tasks including corneal nerve segmentation and DPN diagnosis. 
To facilitate robust self-supervised learning (SSL), we aggregated a large pretraining dataset of 6,426 corneal confocal microscopy (CCM) images from multiple sources. As detailed in Table~\ref{tab:data}, this collection includes the four subsets of the public CORN dataset \cite{imed_ningbo} and the clinical CCM Dataset (UOM CCM) mentioned above. For pretraining, all existing labels were disregarded, and only the raw image data was used to enable the model to learn fundamental feature representations.

\begin{table}[!htbp]
    \centering
    \begin{tabular}{lll}
    \hline
    \textbf{Dataset} & \textbf{Description} & \textbf{Count} \\ \hline
    CORN-1 & Nerve segmentation subset & 1698 \\
    CORN-2 & Image enhancement subset & 688 \\
    CORN-3 & Nerve tortuosity grading & 403 \\
    CORN1500 & Extended tortuosity grading & 1500 \\
    UoM CCM & Diabetes classification dataset & 2137 \\ \hline
    \end{tabular}
    \caption{Datasets aggregated for self-supervised pretraining.}
    \label{tab:data}
\end{table}

The dataset used in this study, provided by the Early Neuropathy Assessment (ENA) group from the University of Manchester, UK, consisted of images of the corneal sub-basal nerve plexus. These images were sourced from healthy volunteers (HV) and participants with either prediabetes or diabetes, encompassing a total of 318 individuals. For each participant, 4 to 8 CCM images were selected based on image quality to ensure optimal performance \cite{kalteniece2017corneal}. The corneal confocal microscopy (CCM) images were obtained using an internationally recognized protocol developed by the ENA group. The imaging was conducted with the Rostock Corneal Module of a Heidelberg Retina Tomograph III confocal laser microscope (HRTII32-RCM, Heidelberg Engineering, Germany) at a resolution of 400×400 $\mu$m (384 × 384 pixels). The images were exported in BMP file format to ensure compatibility with the image analysis software.

The participants were categorised into three groups: 
Healthy Volunteers (HV): 112 patients (35.2\%), no diabetes or neuropathy.
Diabetic Patients without Neuropathy ($\text{PN}-$): 98 patients (30.8\%), diabetes duration $\geq$ 5 years.
Diabetic Patients with Neuropathy ($\text{PN}+$): 108 patients (34.0\%), confirmed DPN symptoms. The demographic and clinical characteristics of the study participants are summarised in Table~\ref{tab:demographics}.

\begin{table}[!htbp]
\centering
\caption{Demographic and clinical characteristics of the study participants. Values are presented as Mean $\pm$ Standard Deviation (SD) or Count.}
\label{tab:demographics}
\resizebox{0.5\textwidth}{!}{%
\begin{tabular}{lccc}
\hline
Variable & Control & with DPN (PN+) & without DPN (PN-) \\ \hline
Duration of diabetes (years) & 0 $\pm$ 0 & 45.54 $\pm$ 11.96 & 23.5 $\pm$ 13.74 \\
Age & 39.64 $\pm$ 14.15 & 60.79 $\pm$ 11.40 & 41.85 $\pm$ 14.46 \\
Gender (Female/Male) & 56.4\%/43.6\% & 52.8\%/47.2\% & 44.6\%/55.4\% \\
BMI ($kg/m^2$) & 25.64 $\pm$ 4.74 & 27.55 $\pm$ 3.90 & 26.22 $\pm$ 4.56 \\
HbA1c & 5.56 $\pm$ 0.34 & 8.27 $\pm$ 1.34 & 8.34 $\pm$ 1.45 \\
IFCC (mmol/mol) & 36.34 $\pm$ 5.51 & 66.60 $\pm$ 14.25 & 67.60 $\pm$ 15.79 \\ \hline
\end{tabular}%
}
\end{table}

\begin{figure}[!htbp]
    \centering
    \includegraphics[width=0.5\textwidth]{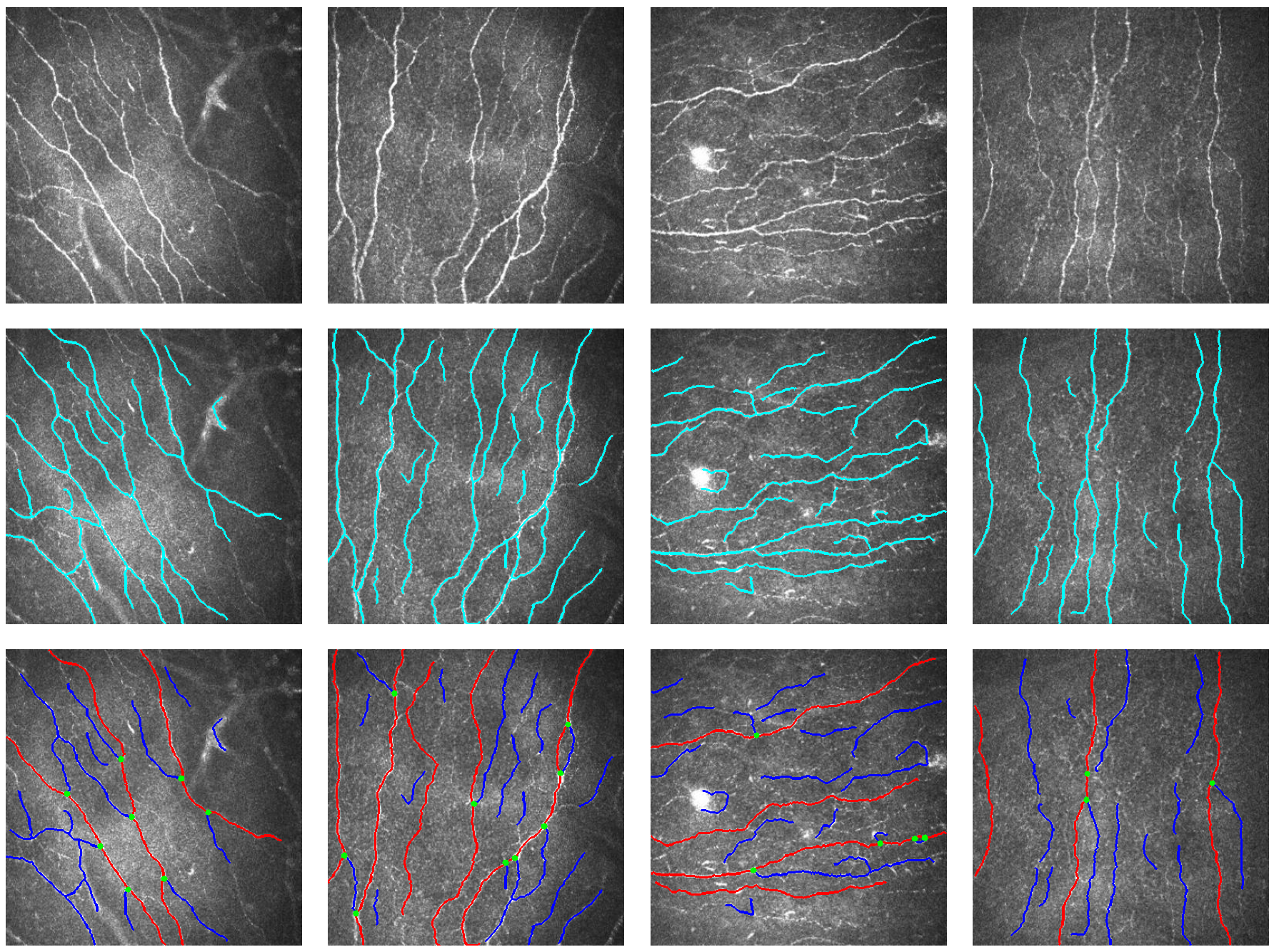}
    \caption{Visualisation of corneal nerve fibre analysis from corneal confocal microscopy (CCM) images. Top row: original corneal nerve images from three subjects. Middle row: ground-truth manual annotations (cyan traces). Bottom row: automated nerve fibre segmentation showing main nerves (red) and branches (blue), with detected branching points (green dots).}
    \label{fig:dataset}
\end{figure}

The manual annotations of the nerve fibre centerlines were traced by an ophthalmologist using the open-source software ImageJ \cite{collins2007imagej}. {Figure~\ref{fig:dataset}} illustrates the corneal nerve fibre analysis, displaying the original CCM images, manual annotations, and automated nerve fibre segmentation analysis. In this analysis, longer nerve fibres with larger diameters are identified as main trunks, while fibres originating from these trunks are classified as nerve branches. This classification, performed using ACCMetrics \cite{dabbah2011automatic}, forms the basis for calculating CNFL and CNBD.

\subsection{Implementation Details}

All experiments were conducted on a workstation equipped with an NVIDIA RTX A6000 GPU and an Intel Xeon W-2255 CPU, using PyTorch 2.1 with CUDA 12 and mixed-precision training. 

\textbf{Self-supervised pretraining:} We employed a block-masked reconstruction strategy with a patch size of $4\times 4$, and blocks composed of $4\times 4$ patches. Sensitivity analysis (ratios 50\%--90\%) identified 75\% as the optimal masking ratio (yielding 85.6\% diagnostic accuracy and 61.34\% mIoU), balancing reconstruction difficulty with semantic retention---lower ratios made tasks too trivial, while higher ratios removed critical context. Models were trained for 400 epochs using the AdamW optimiser (learning rate 1.5e-4) with a cosine decay schedule following a 10-epoch warm-up. 

\textbf{Downstream tasks training:} Models were fine-tuned for 160 epochs using a combined loss function and the same cosine decay schedule.

To ensure robust and unbiased evaluation, we implemented a patient-level 5-fold cross-validation scheme. The 318 participants were randomly partitioned into five folds, with all images from a single participant confined to the same fold to prevent data leakage. In each cross-validation run, three folds were used for training, one for validation, and one for testing. The final reported metrics are the average performance across the five test folds. The source code for this project will be made publicly available at \url{https://gitlab.com/han-research/hmsvit}.

\section{Result}

\subsection{Model Performance Evaluation}

\begin{table*}[!htbp]
    \centering
    \begin{adjustbox}{max width=\textwidth}
        \begin{tabular}{llllllllll}
        \hline
\multirow{2}{*}{Model} & \multirow{2}{*}{Size} & \multicolumn{1}{c}{\multirow{2}{*}{Channels}} & \multirow{2}{*}{Blocks} & \multicolumn{1}{c}{\multirow{2}{*}{Param}} & \multicolumn{2}{c}{Nerve Segmentation}                      & \multicolumn{2}{c}{Diagnostic   Accuracy}       \\
                       &                       & \multicolumn{1}{c}{}                          &                         & \multicolumn{1}{c}{}                       & \multicolumn{1}{c}{mIoU(\%)} & \multicolumn{1}{c}{Accuracy} & \multicolumn{1}{c}{Image-Level} & Patient-Level \\ \hline
Swin                   & Tiny                  & {[}96‑192‑384‑768{]}                          & \multirow{3}{*}{12}     & 29M                                        & 54.45 ± 0.72                 & 0.960 ± 0.005                & 0.623 ± 0.015                   & 0.762 ± 0.013 \\
HiViT                  & Tiny                  & {[}96‑192‑384{]}                              &                         & 20M                                        & 52.80 ± 0.78                 & 0.950 ± 0.006                & 0.613 ± 0.016                   & 0.753 ± 0.014 \\
\textbf{HMSViT}        & Tiny                  & {[}96‑192‑384‑768{]}                          &                         & 28M                                        & 57.39 ± 0.48                 & 0.985 ± 0.003                & 0.643 ± 0.010                   & 0.785 ± 0.009 \\
Swin                   & Small                 & {[}96‑192‑384‑768{]}                          & \multirow{3}{*}{16}     & 50M                                        & 56.10 ± 0.65                 & 0.970 ± 0.005                & 0.643 ± 0.013                   & 0.794 ± 0.012 \\
HiViT                  & Small                 & {[}96‑192‑384{]}                              &                         & 38M                                        & 55.00 ± 0.70                 & 0.965 ± 0.006                & 0.653 ± 0.014                   & 0.802 ± 0.013 \\
\textbf{HMSViT}        & Small                 & {[}96‑192‑384‑768{]}                          &                         & 35M                                        & 58.55 ± 0.45                 & 0.985 ± 0.003                & 0.676 ± 0.009                   & 0.832 ± 0.008 \\
Swin                   & Base                  & {[}128‑256‑512‑1024{]}                        & \multirow{3}{*}{24}     & 88M                                        & 58.30 ± 0.68                 & 0.975 ± 0.005                & 0.684 ± 0.012                   & 0.823 ± 0.010 \\
HiViT                  & Base                  & {[}128‑256‑512{]}                             &                         & 67M                                        & 57.20 ± 0.73                 & 0.970 ± 0.006                & 0.676 ± 0.013                   & 0.817 ± 0.012 \\
\textbf{HMSViT}        & Base                  & {[}128‑256‑384‑512{]}                         &                         & 52M                                        & 61.34 ± 0.40                 & 0.984 ± 0.003                & 0.704 ± 0.008                   & 0.856 ± 0.007 \\ \hline
\end{tabular}
\end{adjustbox}
\caption{Performance comparison of the proposed HMSViT with hierarchical vision transformers Swin \cite{Liu2021Swin} and HiViT \cite{Zhang2022HiViT}. For the Tiny, Small, and Base variants, we report each model’s backbone configuration (channels, blocks) and parameter count (Params), together with performance on corneal nerve segmentation and diabetic neuropathy diagnosis. Values are mean ± SD over 5-fold cross-validation.}
    \label{tab:1}
\end{table*}

\begin{figure*}[!htbp]
    \centering
    \includegraphics[width=\textwidth]{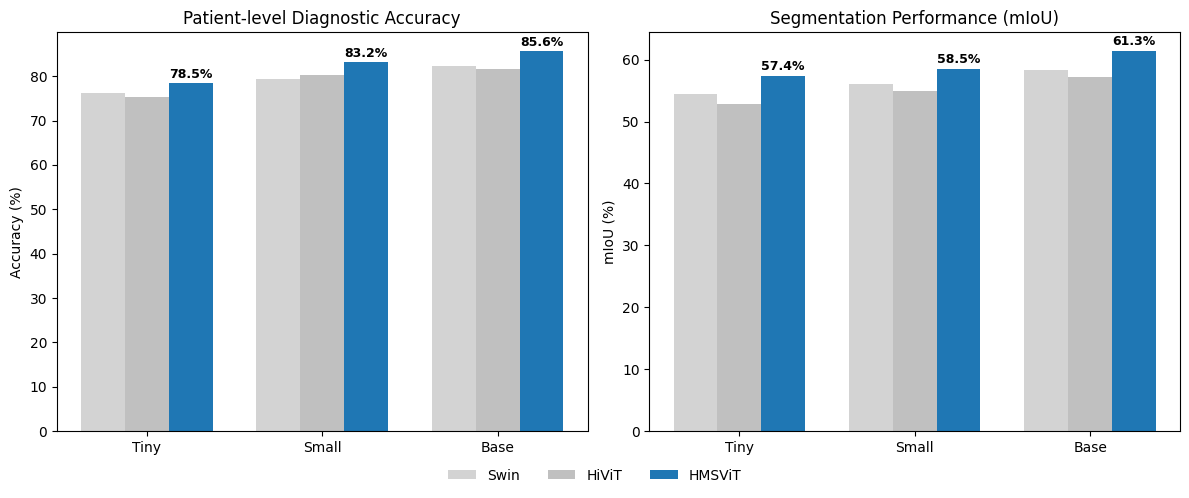}
    \caption{Performance gains of HMSViT over Swin and HiViT across Tiny, Small, and Base configurations in patient-level diagnostic accuracy and segmentation mIoU.}
    \label{fig:perf}
\end{figure*}

\begin{table}[!htbp]
    \centering
    \begin{adjustbox}{max width=0.5\textwidth}
        \begin{tabular}{lccccccccc}
            \hline
            \multirow{2}{*}{Class} & \multicolumn{3}{c}{Tiny} & \multicolumn{3}{c}{Small} & \multicolumn{3}{c}{Base} \\
            \cline{2-10} 
                                   & Precision & Recall & F1-score & Precision & Recall & F1-score & Precision & Recall & F1-score \\ \hline
            HV                      & 0.730      & 0.790   & 0.760     & 0.695     & 0.802  & 0.745    & 0.812     & 0.758  & 0.784    \\
            PN+                     & 0.570      & 0.530   & 0.550     & 0.667     & 0.667  & 0.667    & 0.615     & 0.667  & 0.640    \\
            PN-                     & 0.550      & 0.510   & 0.530     & 0.650     & 0.527  & 0.582    & 0.645     & 0.662  & 0.653    \\ \hline
            \textbf{Accuracy}       & \multicolumn{3}{c}{0.643} & \multicolumn{3}{c}{0.676} & \multicolumn{3}{c}{0.704} \\ \hline
            \end{tabular}
    \end{adjustbox}
    \caption{Detailed image-level diagnostic classification performance of HMSViT across Tiny, Small, and Base configurations. The table presents precision, recall, and F1-score for each class: Healthy Volunteers (HV), Diabetic Patients with Neuropathy (PN+), and Diabetic Patients without Neuropathy (PN-).}
    \label{tab:diagnostic_details}
\end{table}

We evaluated HMSViT on corneal nerve segmentation and diabetic neuropathy (DPN) diagnosis using three configurations (Tiny, Small, Base) and compared its performance with state-of-the-art hierarchical models, Swin Transformer \cite{Liu2021Swin} and HiViT \cite{Zhang2022HiViT}.

\textbf{Overall Performance:} As detailed in Table~\ref{tab:1} and illustrated in Figure~\ref{fig:perf}, HMSViT consistently outperforms both Swin Transformer and HiViT across all configurations in nerve segmentation and diagnostic accuracy. Notably, the HMSViT-Base model achieves a patient-level diagnostic accuracy of 85.6\% and a segmentation mIoU of 61.34\%, surpassing Swin-Base (82.3\% accuracy, 58.30\% mIoU) and HiViT-Base (81.7\% accuracy, 57.20\% mIoU). This performance advantage is achieved with substantially fewer parameters (52M) than Swin (88M) and HiViT (67M), underscoring the efficiency of the proposed pooling-based hierarchical architecture and the block-masked self-supervised pretraining strategy.

\textbf{Diagnostic Classification Performance:} Table~\ref{tab:diagnostic_details} provides a detailed breakdown of the image-level classification performance. HMSViT-Base demonstrates strong and balanced metrics across all classes, achieving an F1-score of 0.784 for Healthy Volunteers (HV), 0.640 for PN+, and 0.653 for PN-. The overall image-level accuracy for HMSViT-Base is 70.40\%, which is higher than Swin-Base (68.42\%) and HiViT-Base (67.55\%) as shown in Table~\ref{tab:1}. This indicates that the features learned through our self-supervised approach are highly discriminative for identifying subtle neuropathic changes.

\textbf{Corneal Nerve Fibre Analysis:} In addition to segmentation and classification, we assessed the model's ability to quantify clinically relevant nerve morphology parameters. Table~\ref{tab:2} shows the Root Mean Square Error (RMSE) and Standard Deviation (SD) for Corneal Nerve Fibre Length (CNFL) and Corneal Nerve Branch Density (CNBD). The HMSViT-Base configuration yields the lowest RMSE (326.002 for CNFL, 4.1603 for CNBD) and SD values, indicating its superior precision in quantifying nerve structures compared to the Tiny and Small variants. This highlights the model's potential for reliable clinical biomarker extraction.

In summary, HMSViT demonstrates robust and superior performance in both segmentation and diagnosis, leveraging efficient multi-scale feature extraction and advanced self-supervised learning to deliver a powerful tool for DPN screening.

\subsection{Ablation Study to Evaluate the Impact of Hierarchical Design and Self-Supervised Learning}

\begin{table}[!htbp]
  \centering
  \begin{minipage}{0.45\textwidth}
    \centering
    \begin{adjustbox}{max width=\textwidth}
        \begin{tabular}{lllll}
            \hline
            \multicolumn{2}{c}{HMSViT}   & Tiny    & Small   & Base    \\ \hline
            \multirow{2}{*}{CNFL} & RMSE & 501.623 & 463.471 & 326.002 \\
                                  & SD   & 500.63  & 302.231 & 271.745 \\ \hline
            \multirow{2}{*}{CNBD} & RMSE & 7.3667  & 5.1326  & 4.1603  \\
                                  & SD   & 6.911   & 4.3934  & 4.1652  \\ \hline
            \end{tabular}
    \end{adjustbox}
    \caption{Performance evaluation of hierarchical multi-scale Vision Transformer (HMSViT) variants (Tiny, Small, Base) on corneal nerve biomarker prediction. The table reports Root Mean Square Error (RMSE) and standard deviation (SD) for Corneal Nerve Fibre Length (CNFL) and Corneal Nerve Branch Density (CNBD). The Base model achieves the lowest error and variability for both biomarkers, indicating improved predictive performance with increased model capacity.}
    \label{tab:2}
  \end{minipage}%
  \hfill
  \begin{minipage}{0.45\textwidth}
    \centering
    \begin{adjustbox}{max width=\textwidth}
            \begin{tabular}{lll}
            \hline
            \multicolumn{3}{c}{Diagnostic classification accuracy}     \\ \hline
            \multicolumn{1}{c}{} & No Hierarchical & With Hierarchical \\ \hline
            Supervised           & 0.656           & 0.666             \\
            Self-supervised      & 0.694           & 0.704             \\ \hline
            \multicolumn{3}{l}{Corneal nerve segmentation performance (\%)} \\ \hline
            Supervised           & 54.31           & 57.28             \\
            Self-supervised      & 55.14           & 61.34             \\ \hline
            \end{tabular}
    \end{adjustbox}
    \caption{Performance comparison of supervised and self-supervised learning approaches for diagnostic classification and corneal nerve segmentation, with and without hierarchical modelling. Self-supervised methods outperform supervised ones across both tasks, with notable improvements when hierarchical modelling is incorporated, particularly in corneal nerve segmentation accuracy.}
    \label{tab:3}
  \end{minipage}
\end{table}

Table~\ref{tab:3} summarised the results of these settings in terms of classification accuracy and segmentation performance.

For DPN diagnosis, the results demonstrated that self-supervised learning (SSL) consistently enhanced model performance compared to supervised training alone. Specifically, the model trained with SSL and a hierarchical structure achieved the highest classification accuracy of 70.40\%, while supervised learning without the hierarchical structure yielded 65.60\% accuracy. This indicates that SSL effectively leverages unlabelled data to improve diagnostic capability. Moreover, incorporating the hierarchical structure provided additional gains, increasing supervised learning accuracy from 65.60\% (without hierarchical design) to 66.60\% (with hierarchical design).

For nerve segmentation, the segmentation quality measured by mean Intersection over Union (mIoU) improved significantly with both the hierarchical structure and SSL. The supervised model without the hierarchical structure achieved an mIoU of 54.31\%, while adding the hierarchical structure increased it to 57.28\%. Notably, the combination of SSL and the hierarchical structure resulted in the highest mIoU of 61.34\%, underscoring the benefits of integrating multi-scale feature extraction with self-supervised learning.

\subsection{Comparison with State-of-the-Art Methods for Segmentation}

\begin{table*}[!htbp]
    \centering
    \begin{adjustbox}{max width=0.9\textwidth}
        \begin{tabular}{cccccc}
            \hline
            Models          & Backbone & Self-supervised & Param & Inf. Time (ms) & mIoU (\%) \\ \hline
            DPT\cite{Ranftl2021Vision}             & ViT      & None            & 97M   & 22.8           & 54.31     \\
            PoolFormer\cite{Yu2022MetaFormer}       & HiViT    & None            & 77M   & 17.1           & 55.32     \\
            SegFormer\cite{Xie2021SegFormer}       & HiViT    & None            & 85M   & 19.5           & 57.39     \\
            MOCOV3\cite{Chen2021Empirical}+ViT      & ViT      & MOCOV3          & 86M   & 20.1           & 56.62     \\
            MAE\cite{He2021Masked} +ViT         & ViT      & MAE             & 86M   & 19.8           & 57.08     \\
            MAE+HiViT\cite{Zhang2022HiViT}       & HiViT    & MAE modified    & 73M   & 16.5           & 61.06     \\
            MixMIM\cite{Liu2023MixMAE}          & HiViT    & MAE modified    & 88M   & 20.7           & 60.86     \\
            Fast-iTPN\cite{Tian2024Fast}       & HiViT    & MAE modified    & 79M   & 17.9           & 61.25     \\
            Proposed method & HMSViT   & MAE modified    & 68M   & 15.2           & 61.34     \\ \hline
            \end{tabular}
    \end{adjustbox}
    \caption{Benchmarking HMSViT against state-of-the-art segmentation models. Our proposed method achieves the highest mIoU with the fewest parameters and fastest inference time, demonstrating a superior balance of accuracy and efficiency.}
    \label{tab:4}
\end{table*}

In Table~\ref{tab:4}, we compared the performance and computational efficiency of state-of-the-art vision transformer-based segmentation models. The DPT model, built on a standard Vision Transformer architecture with 97M parameters, achieved an mIoU of 54.31\% with an inference time of 22.8 ms. In contrast, hierarchical models like PoolFormer (77M parameters, 17.1 ms) and SegFormer (85M parameters, 19.5 ms) showed modest improvements, reaching mIoU scores of 55.32\% and 57.39\%, respectively.

Notably, the application of self-supervised learning further enhanced performance. The models utilising MAE-based techniques, such as MAE+ViT (86M parameters, 19.8 ms, mIoU: 57.08\%) and MAE+HiViT (73M parameters, 16.5 ms, mIoU: 61.06\%), benefited from the integration of a hierarchical design. Similarly, Fast-iTPN, incorporating a modified MAE framework with 79M parameters, achieves an mIoU of 61.25\% in 17.9 ms.

\section{Discussion}

This approach addresses key challenges in medical imaging, where limited labelled data often constrains the effectiveness of deep learning methods. Unlike previous models that rely heavily on large annotated datasets, our self-supervised learning (SSL) strategy enables efficient representation learning from limited labelled data. This is particularly valuable for high-dimensional corneal confocal microscopy (CCM) images, which are often scarce and expensive to annotate. Comparisons with state-of-the-art methods highlight HMSViT’s superior performance over models relying solely on hierarchical ViTs or CNNs.  The proposed model’s multi-scale feature extraction enables the simultaneous capture of fine-grained local details and broader structural context, which is critical for accurate segmentation. This capability not only improves diagnostic reliability but also facilitates the extraction of clinically relevant parameters, such as corneal nerve fibre length (CNFL) and branch density.

\begin{figure}[!htbp]
    \centering
    \includegraphics[width=0.5\textwidth]{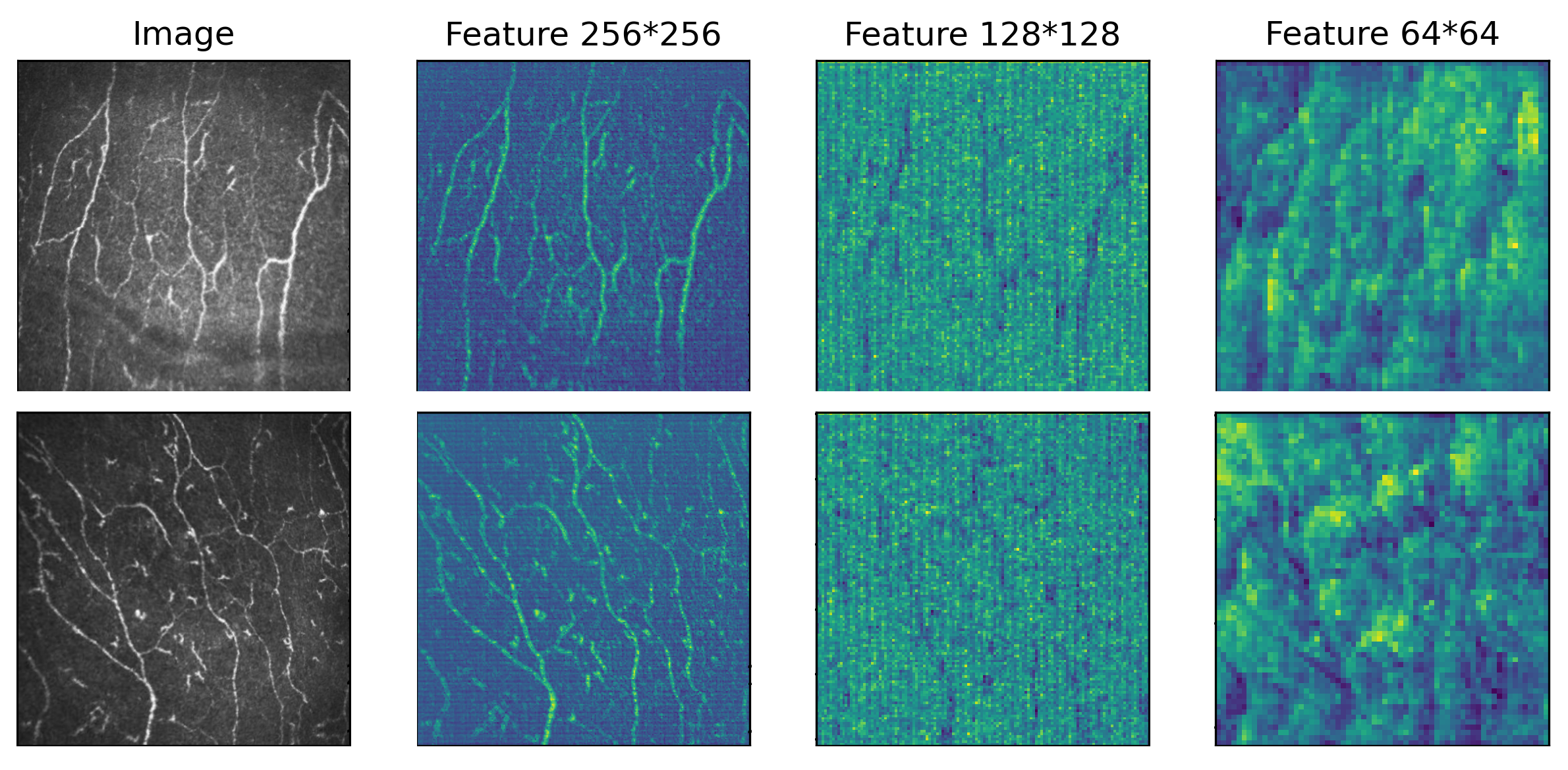}
    \caption{Multi-scale feature representations extracted from a Vision Transformer model applied to corneal confocal microscopy images. The first column shows the original input images. The second, third, and fourth columns illustrate feature maps at different spatial resolutions (256×256, 128×128, and 64×64), capturing hierarchical representations from fine to coarse levels. Fine features highlight detailed nerve structures, while coarser features capture more abstract patterns.}
    \label{fig:feature_maps}
\end{figure}

One of the key innovations of the HMSViT model is its ability to capture hierarchical multi-scale features, effectively combining local and global contextual information. This capability is crucial in medical imaging tasks where both fine-grained details and broader structural patterns contribute to accurate diagnosis. As illustrated in {Figure~\ref{fig:feature_maps}}, the hierarchical feature extraction approach enables the model to enhance segmentation accuracy, as evidenced by its superior performance in mean Intersection over Union (mIoU) compared to existing models. The hierarchical nature of the HMSViT allows for better retention of nerve fibre structure and branching patterns, ensuring more precise morphological assessments.
Another innovation in HMSViT is its streamlined design featuring a pooling-based token mixer, which minimizes architectural complexity while maintaining robust performance. This simplicity is key to the model’s computational efficiency. Notably, HMSViT achieved an mIoU of 61.34\% using only 68M parameters, outperforming more complex architectures that required significantly higher computational resources. Such efficiency is critical for deployment in clinical environments where processing speed and resource constraints are paramount.

A significant challenge in medical imaging is the scarcity of labelled datasets, which limits the performance of supervised deep learning models. The proposed HMSViT mitigates this limitation by leveraging self-supervised learning, which enables the model to learn meaningful representations from large amounts of unlabelled data. The ablation study results confirmed that the addition of self-supervised learning significantly enhanced both segmentation and classification performance. Notably, the model trained with SSL and the hierarchical structure achieved the highest classification accuracy of 70.40\% and an mIoU of 61.34\%. These improvements underscore the importance of SSL in enabling efficient feature learning, reducing reliance on expensive manual annotations.

A direct comparison of the proposed model with other state-of-the-art methods highlights its superior performance. While previous hierarchical vision transformers such as HiViT and Fast-iTPN achieved competitive results, the HMSViT surpassed these models in segmentation performance due to its optimised pyramid structure and modified Masked Autoencoder (MAE) approach. The ability to outperform models that rely solely on hierarchical architectures or traditional convolutional networks demonstrates the effectiveness of integrating SSL with multi-scale feature extraction. The proposed approach not only improves segmentation quality but also enhances diagnostic accuracy, reinforcing its potential for real-world clinical applications.

Furthermore, the ability to quantify clinically relevant parameters such as Corneal Nerve fibre Length (CNFL) and Corneal Nerve Branch Density (CNBD) demonstrates the practical utility of the HMSViT. By extracting these biomarkers with high precision, the model supports objective and reproducible assessments, which are critical for tracking disease progression and evaluating treatment efficacy. 

Despite its advantages, the proposed approach has certain limitations. While HMSViT has shown strong performance on the available dataset, its generalisability to larger and more diverse datasets remains to be validated. Future studies should explore the model’s robustness across different populations, imaging conditions, and disease severities to ensure broad applicability.         
Additionally, although self-supervised learning significantly reduces the need for annotated data, further enhancements in unsupervised learning strategies could be explored to maximize performance. The integration of contrastive learning techniques or multi-modal learning approaches incorporating additional clinical data (e.g., patient history, genetic markers) could further enhance the model’s diagnostic capabilities.

\section{Conclusion}

In this work, we presented HMSViT, a Hierarchical Multi-Scale Vision Transformer that integrates a novel block-masked self-supervised learning strategy with a streamlined hierarchical architecture. By combining a pooling-based hierarchical design with a multi-scale decoder, HMSViT successfully captures both fine-grained local details and broader global contexts from corneal confocal microscopy images. Our extensive experiments on clinical datasets demonstrated that HMSViT achieves state-of-the-art performance. Notably, it attained a patient-level diagnostic accuracy of 85.60\% and a segmentation mIoU of 61.34\%, outperforming leading models like the Swin Transformer and HiViT. This superior accuracy was achieved while using significantly fewer parameters—for instance, approximately 41\% fewer than the Swin Transformer. Ablation studies further confirmed that the synergy between our hierarchical design and block-masked pretraining is critical to its success, driving the substantial improvements in both segmentation and classification tasks.

Looking ahead, future work will focus on validating the model on larger and more diverse datasets to ensure robustness across different imaging conditions and patient populations. We also plan to enhance the model’s interpretability and explore complementary self-supervised techniques—such as contrastive and multimodal learning, to further improve diagnostic performance and facilitate integration into clinical workflows.

\section*{CRediT authorship contribution statement}

\textbf{Xin Zhang:} Conceptualization, Methodology, Software, Validation, Formal analysis, Investigation, Visualization, Writing – original draft. \textbf{Liangxiu Han:} Conceptualization, Methodology, Supervision, Project administration, Funding acquisition, Writing – review \& editing. \textbf{Yue Shi:} Software, Validation, Writing – review \& editing. \textbf{Yalin Zheng:} Data curation, Resources, Project administration, Funding acquisition, Writing – review \& editing. \textbf{Uazman Alam:} Data curation, Resources, Writing – review \& editing. \textbf{Maryam Ferdousi:} Data curation, Resources, Writing – review \& editing. \textbf{Rayaz Malik:} Conceptualization, Data curation, Resources, Writing – review \& editing.

\section*{Acknowledgments}
This work was supported by the Engineering and Physical Sciences Research Council (EPSRC) through grants EP/X013707/1 and EP/X01441X/1.

\section*{Declaration of competing interest}
The authors declare that they have no known competing financial
interests or personal relationships that could have appeared to influence
the work reported in this paper.

\section*{Data Availability}
The datasets generated and analysed during the current study are not publicly available due to patient privacy restrictions but are available from the corresponding author on reasonable request. All studies contributing to the clinical dataset received ethical approval from the NRES Committee North West - Greater Manchester committees, including the NIH study (Ophthalmic Markers of Diabetic Neuropathy, REC number: 09/H1006/38), the LANDMARK study (Surrogate markers for diabetic neuropathy, REC number: 08/H1004/1), and the PROPANE study (Probing the Role of Sodium Channels in Painful Neuropathies, REC number: 14/NW/0093). Written informed consent was obtained from all participants.

\section*{Code Availability}
The source code for this project is publicly available at \url{https://gitlab.com/han-research/hmsvit}.

\bibliographystyle{IEEEtran}
\bibliography{eye}
\end{document}